\documentclass[conference]{IEEEtran}
\IEEEoverridecommandlockouts
\usepackage{cite}
\usepackage{amsmath,amssymb,amsfonts}
\usepackage{array}
\usepackage{booktabs}
\usepackage{graphicx}
\usepackage{textcomp}
\usepackage{xcolor}
\usepackage{url}

\definecolor{chunkbar}{HTML}{6B7280}
\definecolor{entitybar}{HTML}{2563EB}
\definecolor{parentbar}{HTML}{F59E0B}
\definecolor{bmbar}{HTML}{10B981}
\definecolor{softline}{HTML}{E5E7EB}

\def\BibTeX{{\rm B\kern-.05em{\sc i\kern-.025em b}\kern-.08em
    T\kern-.1667em\lower.7ex\hbox{E}\kern-.125emX}}

\begin{document}

\title{EAR: Entity-Aware Partitioning Approach for Retrieval-Augmented Generation Development}

\author{\IEEEauthorblockN{Cenab Batu Bora\IEEEauthorrefmark{1},
Oylum Alatl{\i}\IEEEauthorrefmark{2},
Sebnem Bora\IEEEauthorrefmark{2},
O\u{g}uz Dikenelli\IEEEauthorrefmark{2}}
\IEEEauthorblockA{\IEEEauthorrefmark{1}Georgia Institute of Technology, Atlanta, GA, USA; cenab@gatech.edu\\
\IEEEauthorrefmark{2}Ege University, Izmir, Turkey\\
oylum.alatli@ege.edu.tr; sebnem.bora@ege.edu.tr; oguz.dikenelli@ege.edu.tr}}

\IEEEpubid{\makebox[\columnwidth]{979-8-3195-4709-5/26/\$31.00 ©2026 IEEE\hfill}\hspace{\columnsep}\makebox[\columnwidth]{}}

\maketitle

\begin{abstract}
Retrieval-augmented generation (RAG) can improve knowledge-intensive question answering, but the first design choice is easy to overlook: how should the source corpus be partitioned into retrievable units? Fixed-size chunks often return long passages whose relation to the question is only implicit. We introduce EAR, an Entity-Aware Partitioning approach for multiple-choice question answering (MCQA). EAR extracts normalized surface anchors from the question, answer options, and corpus; retrieves local windows around matching corpus anchors; and can attach a larger parent passage through an extractive summary. We evaluate EAR on a cleaned Massive Multitask Language Understanding (MMLU)-style subset of 153 questions selected by an automatic corpus-support heuristic and using decontaminated public textbook text. Across same-protocol top-$k=3$ and top-$k=8$ sweeps with Mistral, Gemma, and DeepSeek, EAR entity-window reduces retrieved words by 37.5--40.2\% relative to chunks. Observed accuracy changes are +5.2, +1.3, and -3.9 points at top-$k=3$, and +5.9, -3.3, and -4.6 points at top-$k=8$; none of the entity-window differences is statistically significant. The scoped contribution is methodological: EAR provides a compact and inspectable retrieval unit, while its rule-based anchor extractor remains domain-specific and requires separate validation before transfer.
\end{abstract}


\section{Introduction}
Large language models (LLMs) are now routinely evaluated through multiple-choice knowledge benchmarks, including Massive Multitask Language Understanding (MMLU) and localized adaptations of the same evaluation style \cite{hendrycks2021mmlu,yuksel2024turkishmmlu,bayram2025trmmlu}. Retrieval-augmented generation (RAG) can ground such tasks in source documents, but most lightweight systems partition documents before seeing the question: pages become fixed or sliding chunks, and the retriever ranks those pre-made spans \cite{lewis2020rag,gao2024rag}. In the textbook case study used here, the stem and answer options often already name the treaty, ruler, reform, polity, battle, place, or date that should guide retrieval.

EAR builds entity-aware retrieval units by extracting normalized surface anchors from the question and options, matching them to corpus anchors, and retrieving compact windows around those matches. In multiple-choice question answering (MCQA), this makes the retrieval unit responsive to the item rather than only to a document segmentation rule. Generic chunks can mix the answer cue, distractors, and unrelated narrative; longer contexts also increase inference cost and can make evidence use less reliable \cite{liu2024lost}. The central question is therefore direct: what accuracy and context-cost behavior results when generic chunks are replaced with entity-aware windows?

We test EAR in one textbook-grounded curriculum setting. The retrieval architecture is reusable, but the current surface-anchor rules are tailored to the language and subject of this corpus; cross-domain transfer is not established here. We compare EAR against a lexical chunk baseline under the same split, model, prompt family, and cleaned corpus. Our contributions are:
\begin{itemize}
\item We introduce and formalize EAR with an auditable rule-based surface-anchor schema, entity-window retrieval, and an optional parent-expanded variant.
\item We construct a cleaned evaluation protocol that removes benchmark rows flagged by automatic checks and exercise material from the retrieval corpus.
\item We compare no-RAG prompting, chunks, EAR entity-window, and EAR parent-expanded retrieval across two local models and one hosted model on a 153-question supported split.
\item We report accuracy, paired uncertainty, context volume, anchor coverage, computational cost, and window-radius sensitivity.
\end{itemize}

\IEEEpubidadjcol
\section{Related Work}
RAG retrieves external evidence and conditions generation on it \cite{lewis2020rag}. Survey work organizes RAG systems around indexing, retrieval, evidence representation, augmentation, and generation-time integration \cite{gao2024rag}. Many systems index fixed-size chunks and rank them with lexical similarity, dense embeddings, hybrid scoring, or Okapi Best Matching 25 (BM25) \cite{robertson2009bm25}. Fixed chunks are easy to cache, but their boundaries are independent of the question.

Retrieval-unit choice matters because LLMs can fail to use relevant information in long contexts \cite{liu2024lost}; Self-Reflective Retrieval-Augmented Generation (Self-RAG) likewise assesses passage usefulness rather than inserting evidence mechanically \cite{asai2024selfrag}. Clinical entity retrieval demonstrates specialized entity-based selection \cite{lopez2025clear}. Hierarchical and graph systems instead construct broader representations: Recursive Abstractive Processing for Tree-Organized Retrieval (RAPTOR) retrieves tree-organized summaries, while Graph Retrieval-Augmented Generation (GraphRAG) uses entity graphs and community summaries \cite{sarthi2024raptor,edge2024graphrag}. EAR occupies a simpler point in this design space: normalized surface anchors select local windows, and parent context is tested as an optional expansion.

Turkish RAG research now covers general pipelines and educational corpora. Bikmaz et al.\ compare Turkish retrieval and generation models \cite{bikmaz2025turkishrag}; Atag\"un et al.\ evaluate four LLMs on Turkish RAG question answering \cite{atagun2025turkishrag}; and RAGTurk evaluates seven pipeline stages, finding that added generative modules can increase cost without stable gains \cite{kose2026ragturk}. Most closely, \"{O}ner et al.\ compare embeddings, generators, and ensembles over Turkish Ministry of Education textbooks \cite{oner2025textbookrag}. TurkishMMLU and TR-MMLU provide the benchmark setting used here \cite{yuksel2024turkishmmlu,bayram2025trmmlu}. These studies optimize models or pipeline stages; EAR asks whether a deterministic surface-anchor window can serve as a compact retrieval unit for MCQA.

\section{Methods}
\subsection{Task and Evaluation Protocol}
The task is five-way MCQA. Each input contains a question stem and answer options A--E, and the generator must return one option label. Throughout the paper, \textit{chunk} denotes the fixed-passage baseline, \textit{EAR entity-window} the main method, \textit{EAR parent-expanded} the broader-context variant, and \textit{BM25 page} a diagnostic high-recall retriever.

All direct comparisons hold the generator, prompt family, split, retrieval corpus, and top-$k$ setting fixed within a run. The main comparison asks how EAR entity-window differs from chunk retrieval under matched conditions. EAR parent-expanded tests broader context, while BM25 page appears only in the Mistral diagnostic.

\subsection{Data Cleaning and Corpus Construction}
The starting benchmark was audited so the experiment would measure retrieval-unit behavior rather than extraction artifacts. The audit produced provisional clean splits, which retain rows marked for review, and strict splits, which keep only rows that pass the automatic audit. The strict test split contains 672 questions. Because the public corpus does not contain evidence for every item, an automatic support heuristic retains a question when the gold option overlaps with enough stem terms in the same retained corpus block. This yields the 153-question supported split. The heuristic controls corpus mismatch but has not been validated against completed human evidence judgments.

The retrieval corpus is built from four public curriculum textbook clean-text files: grade 9, grade 10, grade 11, and grade 12 material. The decontamination pass removes exercise material and multiple-choice blocks before indexing. The final corpus audit reports 1108 pages, 772 pages with retained narrative text, 3955 kept blocks, and 4429 dropped blocks. Table~\ref{tab:data} summarizes the evaluated data regime.

\begin{table}[t]
\caption{Cleaned evaluation and corpus summary.}
\label{tab:data}
\centering
\begin{tabular}{@{}lr@{}}
\toprule
Quantity & Value \\
\midrule
Strict test questions & 672 \\
Supported questions & 153 \\
Supported rate & 0.2277 \\
Usable textbook source files & 4 \\
Total corpus pages & 1108 \\
Pages with narrative text & 772 \\
Kept corpus blocks & 3955 \\
Dropped exercise/question blocks & 4429 \\
\bottomrule
\end{tabular}
\end{table}

\subsection{Proposed Method: EAR}
EAR uses a deterministic surface-anchor schema. Here, \textit{entity} is an operational term for a normalized textual anchor, not the output of a general-purpose named-entity recognition (NER) model. The prose follows Fig.~\ref{fig:ear_algorithms}: Algorithm~1 defines entity-window retrieval and Algorithm~2 the parent-expanded variant. The same extractor handles corpus text (Alg.~1, Step~3) and queries (Step~5). It extracts years, capitalized phrases, and event-like phrases marked by suffixes such as \textit{Sava\c{s}\i}, \textit{Antla\c{s}mas\i}, and \textit{Ferman\i}; answer options with at most six content tokens are also retained as anchors. Normalization handles locale-aware case folding, punctuation variants, numerals, whitespace, and domain suffixes.

EAR requires neither an ontology nor a learned linker or embedding model. Algorithm~1, Steps~3--6 treat surface anchors as auditable retrieval keys and score windows by lexical overlap plus three points per matched anchor and a 1.5-point bonus per matched year/date. If no query anchor matches the index, Step~6 applies lexical scoring over all windows. Ties retain corpus-index order, and Step~7 removes repeated windows.

The corpus-side pipeline cleans pages and removes exercise blocks (Alg.~1, Step~1), segments them into 260-word parent chunks with 60-word overlap (Step~2), extracts corpus anchors (Step~3), and stores a child window extending up to 80 words on each side of an anchor (Step~4). At inference, EAR extracts anchors from the question and options (Step~5), scores candidate windows (Step~6), and returns the top-$k$ deduplicated windows (Step~7).

EAR parent-expanded reuses Algorithm~1 output (Alg.~2, Step~1), retrieves each parent chunk (Step~2), ranks its sentences by query overlap (Step~3), and selects at most two sentences for an extractive summary (Step~4). Step~5 attaches the child, parent text, summary, and anchor-link log before Step~6 removes duplicate parents.

On the supported split, Step~5 extracts at least one question-plus-options anchor for all 153 items and finds at least one exact normalized corpus-index match for 151 (98.7\%). At anchor level, 676 of 1331 extracted anchors match an index key (50.8\%). These values measure operational coverage, not NER precision or recall: the study has no manually labeled entity set.

\begin{figure*}[t]
\centering
\small
\begin{tabular}{@{}p{0.48\textwidth}p{0.48\textwidth}@{}}
\fbox{\begin{minipage}{0.94\linewidth}
\textbf{Algorithm 1: EAR entity-window retrieval}\\
\textbf{Input:} corpus $C$, question $q$, options $O$, top-$k$ budget $k$\\
\textbf{Output:} ranked evidence windows $W_k$\\
1. Clean corpus pages and remove exercise/question blocks.\\
2. Segment each page into overlapping parent chunks.\\
3. Extract normalized corpus entities from each parent chunk.\\
4. Store a child window of up to 80 words on each side of every anchor.\\
5. Extract normalized query entities from $q$ and $O$.\\
6. Score child windows by entity match, lexical overlap, and date match.\\
7. Return the top-$k$ deduplicated windows as evidence.
\end{minipage}}
&
\fbox{\begin{minipage}{0.94\linewidth}
\textbf{Algorithm 2: EAR parent-expanded retrieval}\\
\textbf{Input:} child windows $W_k$, parent chunks $P$, query $q,O$\\
\textbf{Output:} parent-expanded evidence $E_k$\\
1. Run Algorithm 1 to obtain entity-window child windows.\\
2. For each child window, retrieve its parent chunk.\\
3. Rank parent sentences by overlap with the question and options.\\
4. Build a short extractive parent summary from the best sentences.\\
5. Attach the child window, parent summary, and entity-link log.\\
6. Deduplicate repeated parent contexts.\\
7. Return the expanded contexts for generator answering.
\end{minipage}}
\end{tabular}
\caption{Formal EAR algorithms. Algorithm 1 is the entity-window method. Algorithm 2 is the parent-expanded variant.}
\label{fig:ear_algorithms}
\end{figure*}

\subsection{Design Rationale and Scope}
EAR tests a narrow design claim: a useful MCQA retrieval unit can be the local evidence surrounding a question-relevant anchor rather than a uniformly sized passage. The chunk baseline ranks passages against the whole query. EAR first identifies searchable anchors (Fig.~\ref{fig:ear_algorithms}, Alg.~1, Step~5), then ranks local windows around corpus matches (Steps~6--7).

Each context can be traced to a query anchor, corpus anchor, and window through Steps~3--7. A wrong answer can therefore be inspected as missing query extraction at Step~5, poor retrieval at Steps~6--7, or generator failure after relevant evidence was returned.

The retrieval architecture can accept another anchor extractor, but the present capitalization, suffix, and normalization rules are specific to this corpus. Transfer therefore requires redesigning and validating Steps~3 and~5; this study does not establish cross-domain generalization.

\subsection{Baselines, Prompts, and Answering Protocol}
The chunk baseline splits cleaned text into overlapping 220-word chunks with 30-word overlap and ranks them by lexical token overlap against the question plus options. It skips entity indexing and child-window construction in Fig.~\ref{fig:ear_algorithms}, Alg.~1, Steps~3--4. BM25 page retrieval indexes cleaned pages and applies query-aware evidence compression; it is only a diagnostic comparison.

Figure~\ref{fig:prompts} gives the prompt families. Direct rows test zero-shot, few-shot, and chain-of-thought (CoT) prompting without retrieval. Retrieval rows use RAG+CoT: the generator sees context, the question, five shuffled options, and a one-label constraint. Rationales are retained for inspection; scoring uses the parsed A--E answer. The sweep uses Ollama mistral:latest (immutable local model identifier 6577803aa9a0, 7.2B, Q4\_K\_M), Ollama gemma:7b-instruct (identifier a72c7f4d0a15, 9B, Q4\_0), and OpenRouter deepseek/deepseek-v3.2. Each runs at top-$k=3$ and 8 with temperature 0, seed 13, option shuffling, and one generation per item.

\begin{figure*}[t]
\centering
\small
\fbox{\begin{minipage}{0.94\textwidth}
\textbf{Prompt families}

\vspace{0.25em}
\renewcommand{\arraystretch}{1.13}
\begin{tabular}{@{}>{\bfseries}p{0.14\linewidth}@{\hspace{0.9em}}p{0.76\linewidth}@{}}
Short & Question and five options; return one JavaScript Object Notation (JSON) object with a single answer label. \\
Few-shot & The same direct-answer instruction preceded by solved examples. \\
CoT & Question and options; return one answer label plus a brief rationale. \\
RAG & Retrieved context appears before the question and options; answer only from the supplied context when possible. \\
RAG+CoT & Retrieved context plus question and options; return one answer label and a short evidence-based rationale. \\
\end{tabular}

\vspace{0.65em}
\textbf{Main template skeleton}

\vspace{0.25em}
\begin{tabular}{@{}>{\itshape}p{0.14\linewidth}@{\hspace{0.9em}}p{0.76\linewidth}@{}}
Context & Retrieved textbook passages. \\
Question & Original question stem. \\
Options & Shuffled A-E choices. \\
Instruction & Choose exactly one label, use the context when relevant, and return a compact JSON object with an answer field and a brief rationale field. \\
\end{tabular}
\end{minipage}}
\caption{Prompt families. The main EAR comparison uses RAG+CoT; the other templates define the direct, few-shot, and non-rationale baselines for same-protocol sweeps.}
\label{fig:prompts}
\end{figure*}

\subsection{Metrics and Paired Tests}
The primary evaluation metric is end-to-end accuracy on the supported split. Secondary metrics are invalid-answer rate, retrieved words, generation and retrieval latency, and paired accuracy differences. The parser maps each output to A--E and marks malformed outputs invalid. We use a two-sided exact McNemar test on paired question-level correctness and a 2000-resample percentile bootstrap with seed 13 for 95\% confidence intervals (CIs).

Because the experiment requires source retrieval, five-way option selection, corpus-supported evidence, and strict answer parsing, the absolute scores should be interpreted as end-to-end retrieval QA scores rather than closed-book model knowledge scores. The relevant comparison is therefore the accuracy and retrieved-context length of each retrieval unit under the same protocol.

The reported model runs are complete passes over all 153 items. Separate computational diagnostics use an Apple M4 with 32~GB memory and Python 3.11.13. Index build and retrieval latency are means of three runs; peak Python allocation is measured during index construction. A retrieval-only sweep compares 40-, 80-, and 120-word radii at both budgets. Gold-option surface-form presence is logged independently from correctness; it is a cue diagnostic, not proof that a passage supports the answer.

\section{Results}
\subsection{Finding 1: Context Savings Are Consistent; Accuracy Effects Vary}
Table~\ref{tab:main_results} reports the retrieval sweep and Table~\ref{tab:direct_baselines} the matched no-RAG baselines. EAR entity-window uses 37.5\% fewer words than chunks at top-$k=3$ (419.6 versus 671.3) and 40.2\% fewer at top-$k=8$ (1062.3 versus 1775.6).

Accuracy point estimates vary with direct-baseline strength and budget. Mistral's entity-window estimate is above chunks at both budgets (+5.2 and +5.9 points); Gemma is +1.3 at top-$k=3$ and -3.3 at top-$k=8$; DeepSeek is -3.9 and -4.6. Mistral entity-window also lies 2.0--3.3 points above its best direct prompt, while Gemma's highest observed retrieval score is chunk top-$k=8$ (51.0\%). DeepSeek reaches 83.0\% directly, and no retrieval row exceeds it. Thus, retrieval does not improve observed accuracy in this high direct-baseline regime.

Every entity-window CI crosses zero, and across all 12 entity and parent comparisons no result remains significant after Holm or Bonferroni correction (Table~\ref{tab:significance}). The evidence therefore establishes context reduction, not general accuracy superiority. Among retrieval runs, all outputs were valid except one Mistral entity-window output at top-$k=3$ (0.65\%).

\begin{table*}[t]
\caption{Same-protocol retrieval results on the 153-question supported split. Accuracy values are percentages. $\Delta$ is EAR entity-window minus chunk; CIs are paired bootstrap intervals and $p$ is the two-sided exact McNemar value.}
\label{tab:main_results}
\centering
\scriptsize
\begin{tabular}{@{}lccccccc@{}}
\toprule
Model & $k$ & Chunk & EAR entity & EAR parent & $\Delta$ [95\% CI] & $p$ & Entity words \\
\midrule
Mistral & 3 & 37.9 & 43.1 & 45.8 & +5.2 [-3.9, 14.4] & .302 & 419.6 \\
Mistral & 8 & 38.6 & 44.4 & 39.2 & +5.9 [-2.0, 13.7] & .188 & 1062.3 \\
Gemma & 3 & 44.4 & 45.8 & 49.7 & +1.3 [-6.5, 9.2] & .875 & 419.6 \\
Gemma & 8 & 51.0 & 47.7 & 42.5 & -3.3 [-10.5, 3.9] & .487 & 1062.3 \\
DeepSeek & 3 & 80.4 & 76.5 & 79.1 & -3.9 [-9.8, 2.0] & .286 & 419.6 \\
DeepSeek & 8 & 81.0 & 76.5 & 79.1 & -4.6 [-9.8, 0.7] & .167 & 1062.3 \\
\bottomrule
\end{tabular}
\end{table*}

\begin{table}[t]
\caption{No-RAG direct prompt baselines on the same 153-question split with option shuffling and temperature 0. Values are accuracy percentages.}
\label{tab:direct_baselines}
\centering
\scriptsize
\begin{tabular}{@{}lcccc@{}}
\toprule
Model & Zero-shot & Few-shot & CoT & Best \\
\midrule
Mistral & 36.6 & 41.2 & 33.3 & 41.2 \\
Gemma & 38.6 & 34.0 & 39.2 & 39.2 \\
DeepSeek & 83.0 & 79.1 & 82.3 & 83.0 \\
\bottomrule
\end{tabular}
\end{table}

\begin{table*}[t]
\centering
\scriptsize
\caption{Paired tests against chunk retrieval. $\Delta$ is method minus chunk accuracy in points; $p$ is the two-sided exact McNemar value. A dagger marks nominal uncorrected $p<.05$; no row remains significant after Holm or Bonferroni correction across 12 comparisons.}
\label{tab:significance}
\begin{tabular}{@{}lcclcc@{}}
\toprule
Model & $k$ & Method & $\Delta$ [95\% CI] & $p$ & Sig. \\
\midrule
Mistral & 3 & EAR entity & +5.2 [-3.9, 14.4] & .302 & -- \\
Mistral & 3 & EAR parent & +7.8 [-1.3, 16.3] & .104 & -- \\
Mistral & 8 & EAR entity & +5.9 [-2.0, 13.7] & .188 & -- \\
Mistral & 8 & EAR parent & +0.7 [-7.8, 8.5] & 1.000 & -- \\
Gemma & 3 & EAR entity & +1.3 [-6.5, 9.2] & .875 & -- \\
Gemma & 3 & EAR parent & +5.2 [-2.0, 12.4] & .256 & -- \\
Gemma & 8 & EAR entity & -3.3 [-10.5, 3.9] & .487 & -- \\
Gemma & 8 & EAR parent & -8.5 [-15.7, -1.3] & .029 & $\dagger$ \\
DeepSeek & 3 & EAR entity & -3.9 [-9.8, 2.0] & .286 & -- \\
DeepSeek & 3 & EAR parent & -1.3 [-7.2, 4.6] & .824 & -- \\
DeepSeek & 8 & EAR entity & -4.6 [-9.8, 0.7] & .167 & -- \\
DeepSeek & 8 & EAR parent & -2.0 [-7.8, 3.3] & .648 & -- \\
\bottomrule
\end{tabular}
\end{table*}

\subsection{Finding 2: Compact Windows Shift Offline and Online Cost}
Table~\ref{tab:cost_radius} separates offline indexing from online retrieval. EAR constructs 33,228 overlapping windows, so its index takes 6.877~s and 143.0~MiB of peak traced Python allocation, versus 0.011~s and 3.4~MiB for 1397 chunks. Once built, anchor-filtered entity retrieval is faster than exhaustive chunk scoring (24.4--25.2 versus 60.6--62.4~ms per item). Parent expansion is slowest at about 149~ms because Algorithm~2 ranks parent sentences online. Logged end-to-end generation latency is lower for entity-window than chunks in all six model-budget runs, by 1.8--31.4\%; this descriptive range includes local inference and hosted-request latency.

The reported 80-word radius is a fixed engineering choice, not a value tuned for answer accuracy. In the retrieval-only sensitivity check, increasing the radius from 40 to 80 raises the top-$k=3$ gold-option surface-cue rate from .529 to .608; 120 words gives .601 while retrieving 35.5\% more text than 80. At top-$k=8$, 120 reaches .706 versus .673 for 80, but retrieves 1409.8 rather than 1062.3 words. The 80-word radius therefore occupies the middle of the observed coverage--context tradeoff; end-to-end model sweeps at other radii remain future work.

\begin{table*}[t]
\caption{Computational cost and retrieval-only radius sensitivity on 153 questions. Build and peak allocation are index-construction measurements; MiB denotes mebibytes. Online values are mean retrieval milliseconds per question. Cue is gold-option surface-form presence, not human-validated evidence recall.}
\label{tab:cost_radius}
\centering
\scriptsize
\begin{minipage}{0.49\textwidth}
\centering
\textit{(a) Canonical index and online retrieval cost}\\[1pt]
\begin{tabular}{@{}lrrrr@{}}
\toprule
Mode & Units & Build s & Peak MiB & $k=3/8$ ms \\
\midrule
Chunk & 1397 & .011 & 3.4 & 60.6 / 62.4 \\
EAR entity & 33228 & 6.877 & 143.0 & 24.4 / 25.2 \\
EAR parent & same index & -- & -- & 149.3 / 148.6 \\
\bottomrule
\end{tabular}
\end{minipage}\hfill
\begin{minipage}{0.49\textwidth}
\centering
\textit{(b) Entity-window radius sensitivity}\\[1pt]
\begin{tabular}{@{}rrrrr@{}}
\toprule
Radius & Cue $k=3$ & Words $k=3$ & Cue $k=8$ & Words $k=8$ \\
\midrule
40 & .529 & 238.9 & .627 & 597.0 \\
80 & .608 & 419.6 & .673 & 1062.3 \\
120 & .601 & 568.5 & .706 & 1409.8 \\
\bottomrule
\end{tabular}
\end{minipage}
\end{table*}

\subsection{Finding 3: Parent Expansion and Surface Cues Have Limits}
Algorithm~2 parent expansion has the highest observed Mistral and Gemma accuracy at top-$k=3$, but at top-$k=8$ it falls below entity-window for Mistral and below chunks for Gemma and DeepSeek. The added narrative may reintroduce distractors, but this mechanism was not directly tested. At nominal uncorrected $p<.05$, only Gemma top-$k=8$ parent versus chunk differs, negatively: -8.5 points, 95\% CI [-15.7, -1.3], $p=.029$. It does not survive multiple-comparison correction.

Returning a surface cue after Algorithm~1, Step~7 is also not equivalent to answering. In the Mistral top-$k=8$ diagnostic, BM25 page has the highest gold-option surface-cue rate (.882), ahead of chunks (.706), EAR entity-window (.673), and parent expansion (.725), yet its accuracy (.431) is below entity-window (.444). For entity-window, 59 items are correct with a cue present, 9 correct without one, 44 wrong despite one, and 41 wrong without one. These counts make retrieval failure inspectable, but a human evidence annotation is needed to determine whether each cue is genuinely supportive.

\section{Discussion}
EAR changes both context volume and computational placement. Algorithm~1, Steps~3--7 reduce retrieved text by 37.5--40.2\% and move work into a larger offline index; at query time, anchor filtering is faster than exhaustive chunk scoring. Parent expansion reverses that online advantage. This pattern agrees with Turkish RAG evidence that pipeline complexity should be evaluated against cost rather than assumed beneficial \cite{kose2026ragturk}.

Accuracy remains conditional. Mistral has higher entity-window point estimates at both budgets, Gemma changes direction with $k$, and DeepSeek's strong direct baseline is not improved by retrieval. None of the entity-window differences is statistically significant. The modest local-model scores reflect a complete pipeline that retrieves from a restricted corpus, parses one of five shuffled options, and depends on the generator using the returned text; they should not be read as closed-book knowledge estimates.

Evidence access and evidence use also remain separate. BM25 pages contain the gold-option surface form more often than EAR windows in the Mistral diagnostic without producing higher accuracy. EAR localizes possible failures to query-anchor extraction (Fig.~\ref{fig:ear_algorithms}, Alg.~1, Step~5), ranking (Step~6), returned context (Step~7), or generator use. The logs make these stages visible, but only human passage judgments can distinguish genuine supporting evidence from incidental answer-string presence.

The method is therefore a partitioning component, not a complete QA architecture. Relation-aware expansion could invoke Algorithm~2 only for questions about cause, chronology, institutional role, or comparison. A broader evaluation should also compare learned NER and embedding-based extractors, validate the supported split, and rerun the radius sweep end to end.

\section{Limitations}
Four limitations bound the claims. First, the automatic supported split lacks completed human evidence validation. Second, no manually labeled anchor set is available, so the rule-based extractor has coverage diagnostics but no precision, recall, or F1 harmonic mean; its capitalization and suffix rules are domain-specific and can miss aliases, morphology, and cross-sentence references. Third, the radius analysis measures retrieval cues and context length, not answer accuracy. Fourth, 153 questions and three generators do not establish broad model or domain trends. The cost measurements are also hardware-specific.

\section{Conclusion}
We introduced EAR, an Entity-Aware Partitioning approach for RAG development in MCQA. On a 153-question supported split, Algorithm~1 retrieves 37.5--40.2\% fewer words than chunks and reduces measured online retrieval and generation latency, at the cost of a larger offline index. Accuracy point estimates are positive for Mistral, mixed for Gemma, and negative for DeepSeek; no entity-window difference is statistically significant. Parent expansion is not a reliable default. EAR is therefore supported as a compact, inspectable retrieval representation, while extractor validity, corpus support, and cross-domain transfer remain open.

\IEEEtriggeratref{8}


\begin{thebibliography}{00}
\bibitem{hendrycks2021mmlu} D. Hendrycks, C. Burns, S. Basart, A. Zou, M. Mazeika, D. Song, and J. Steinhardt, ``Measuring massive multitask language understanding,'' in \textit{Proc. Int. Conf. Learning Representations}, 2021.
\bibitem{yuksel2024turkishmmlu} A. Y\"uksel, A. K\"oksal, L. K. \c{S}enel, A. Korhonen, and H. Sch\"utze, ``TurkishMMLU: Measuring massive multitask language understanding in Turkish,'' in \textit{Findings of the Association for Computational Linguistics: EMNLP}, 2024.
\bibitem{bayram2025trmmlu} M. A. Bayram, A. A. Fincan, A. S. G\"um\"u\c{s}, B. Diri, S. Y\i ld\i r\i m, and \"O. Ayta\c{s}, ``Setting standards in Turkish NLP: TR-MMLU for large language model evaluation,'' arXiv:2501.00593, 2025.
\bibitem{lewis2020rag} P. Lewis, E. Perez, A. Piktus, F. Petroni, V. Karpukhin, N. Goyal, H. K\"uttler, M. Lewis, W. Yih, T. Rockt\"aschel, S. Riedel, and D. Kiela, ``Retrieval-augmented generation for knowledge-intensive NLP tasks,'' in \textit{Advances in Neural Information Processing Systems}, vol. 33, pp. 9459-9474, 2020.
\bibitem{gao2024rag} Y. Gao \textit{et al.}, ``Retrieval-augmented generation for large language models: A survey,'' arXiv:2312.10997, 2024.
\bibitem{robertson2009bm25} S. Robertson and H. Zaragoza, ``The probabilistic relevance framework: BM25 and beyond,'' \textit{Foundations and Trends in Information Retrieval}, vol. 3, no. 4, pp. 333-389, 2009.
\bibitem{liu2024lost} N. F. Liu \textit{et al.}, ``Lost in the middle: How language models use long contexts,'' \textit{Transactions of the Association for Computational Linguistics}, vol. 12, pp. 157-173, 2024.
\bibitem{lopez2025clear} I. Lopez \textit{et al.}, ``Clinical entity augmented retrieval for clinical information extraction,'' \textit{npj Digital Medicine}, vol. 8, article 45, 2025, doi: 10.1038/s41746-024-01377-1.
\bibitem{asai2024selfrag} A. Asai, Z. Wu, Y. Wang, A. Sil, and H. Hajishirzi, ``Self-RAG: Learning to retrieve, generate, and critique through self-reflection,'' in \textit{Proc. Int. Conf. Learning Representations}, 2024.
\bibitem{sarthi2024raptor} P. Sarthi, S. Abdullah, A. Tuli, S. Khanna, A. Goldie, and C. D. Manning, ``RAPTOR: Recursive abstractive processing for tree-organized retrieval,'' in \textit{Proc. Int. Conf. Learning Representations}, 2024.
\bibitem{edge2024graphrag} D. Edge \textit{et al.}, ``From local to global: A graph RAG approach to query-focused summarization,'' arXiv:2404.16130, 2024.
\bibitem{bikmaz2025turkishrag} E. Bikmaz, M. Briman, and S. Arslan, ``Bridging the language gap in RAG: A case study on Turkish retrieval and generation,'' \textit{Researcher}, vol. 5, no. 1, pp. 38--49, 2025.
\bibitem{atagun2025turkishrag} E. Atag\"un, M. G\"ull\"u, S. Biro\u{g}ul, and N. Bar\i\c{s}\c{c}\i, ``Retrieval-augmented generation in Turkish natural language understanding: A comparative study of large language models,'' \textit{Mu\u{g}la Journal of Science and Technology}, vol. 11, no. 2, pp. 56--65, 2025.
\bibitem{kose2026ragturk} S. K. K\"ose, M. C. Baytekin, B. Akta\c{s}, B. K. G\"or\"ur, E. A. Munis, D. Y\i lmaz, M. Y. Kartal, and C. Toraman, ``RAGTurk: Best practices for retrieval augmented generation in Turkish,'' in \textit{Proc. Second Workshop on Natural Language Processing for Turkic Languages}, pp. 179--196, 2026.
\bibitem{oner2025textbookrag} E. N. \"{O}ner, S. Ceyhun, M. H. Y\i ld\i z, A. Goncharova, T. S. Y\"ucel, H. T. Kesgin, and M. F. Amasyal\i, ``Optimal RAG system design for Turkish textbooks: A comprehensive evaluation and performance enhancement study,'' in \textit{Proc. Innovations in Intelligent Systems and Applications Conf.}, pp. 1--6, 2025, doi: 10.1109/ASYU67174.2025.11208464.
\end{thebibliography}
\end{document}